\documentclass{article}

\usepackage[final,nonatbib]{neurips_2021}
\usepackage[utf8]{inputenc} % allow utf-8 input
\usepackage[T1]{fontenc}    % use 8-bit T1 fonts
\usepackage{hyperref}       % hyperlinks
\usepackage{url}            % simple URL typesetting
\usepackage{booktabs}       % professional-quality tables
\usepackage{amsfonts}       % blackboard math symbols
\usepackage{nicefrac}       % compact symbols for 1/2, etc.
\usepackage{microtype}      % microtypography
\usepackage{amsmath}
\usepackage{amssymb}
\usepackage{booktabs}
\usepackage{upgreek}
\usepackage{graphicx}
\usepackage{caption}
\usepackage{subcaption}
\usepackage{multirow}
\usepackage{enumitem}
\usepackage[ruled,vlined,linesnumbered]{algorithm2e}
\usepackage{xcolor}
\usepackage{multirow}
\usepackage{stackengine}

\definecolor{Gaussian_blue}{RGB}{69, 125, 186}
\definecolor{Gaussian_red}{RGB}{205,86,89}
\definecolor{Gaussian_purple}{RGB}{111,80,175}
\definecolor{ao_english}{rgb}{0.0, 0.5, 0.0}
\def\delequal{\mathrel{\ensurestackMath{\stackon[1pt]{=}{\scriptstyle\Delta}}}}
\newcommand{\fullmodel}{Probabilistic Entity Representation Model}
\newcommand{\model}{PERM}

\let\oldnl\nl% Store \nl in \oldnl
\newcommand{\nonl}{\renewcommand{\nl}{\let\nl\oldnl}}

\title{{\fullmodel} for Reasoning over Knowledge Graphs}
\author{Nurendra Choudhary$^1$, Nikhil Rao$^2$, Sumeet Katariya$^2$, Karthik Subbian$^2$, Chandan K. Reddy$^{1,2}$\\
$^1$Department of Computer Science, Virginia Tech, Arlington, VA\\
$^2$Amazon, Palo Alto, CA\\
\texttt{nurendra@vt.edu, \{nikhilsr, katsumee, ksubbian\}@amazon.com, reddy@cs.vt.edu}}
\begin{document}

\maketitle

\begin{abstract}
%\nr{make sure you're using the 2021 style files. The footer says 2020}
Logical reasoning over Knowledge Graphs (KGs) is a fundamental {technique that can provide efficient querying mechanism %challenge that limits its application 
over large and incomplete databases}. Current approaches employ spatial geometries such as boxes to learn query representations that encompass the answer entities and model the logical operations of projection and intersection. However, their geometry is restrictive and leads to non-smooth strict boundaries, which further results in ambiguous answer entities.
    Furthermore, previous works propose transformation tricks to handle unions which results in non-closure and, thus, cannot be chained in a stream. In this paper, we propose a {\fullmodel} (\model) to encode entities as a Multivariate Gaussian density with mean and covariance parameters to capture its semantic position and smooth decision boundary, respectively. Additionally, we also define the closed logical operations of projection, intersection, and union that can be {aggregated} using an end-to-end objective function. On the logical query reasoning problem, we demonstrate that the proposed {\model} significantly outperforms the state-of-the-art methods on various public benchmark KG datasets on standard evaluation metrics. %by 6.2\% in HITS@3 and 12.6\% in MRR. 
    We also evaluate {\model}'s competence on a COVID-19 drug-repurposing case study and show that our proposed work is able to recommend drugs with substantially better F1 than current methods. Finally, we demonstrate the working of our {\model}'s query answering process through a low-dimensional visualization of the Gaussian representations.%embeddings.% through the query process.
\end{abstract}

\section{Introduction}
\begin{figure*}[htbp]
     \centering
     \begin{subfigure}[b]{0.52\linewidth}
         \centering
         \includegraphics[width=\linewidth]{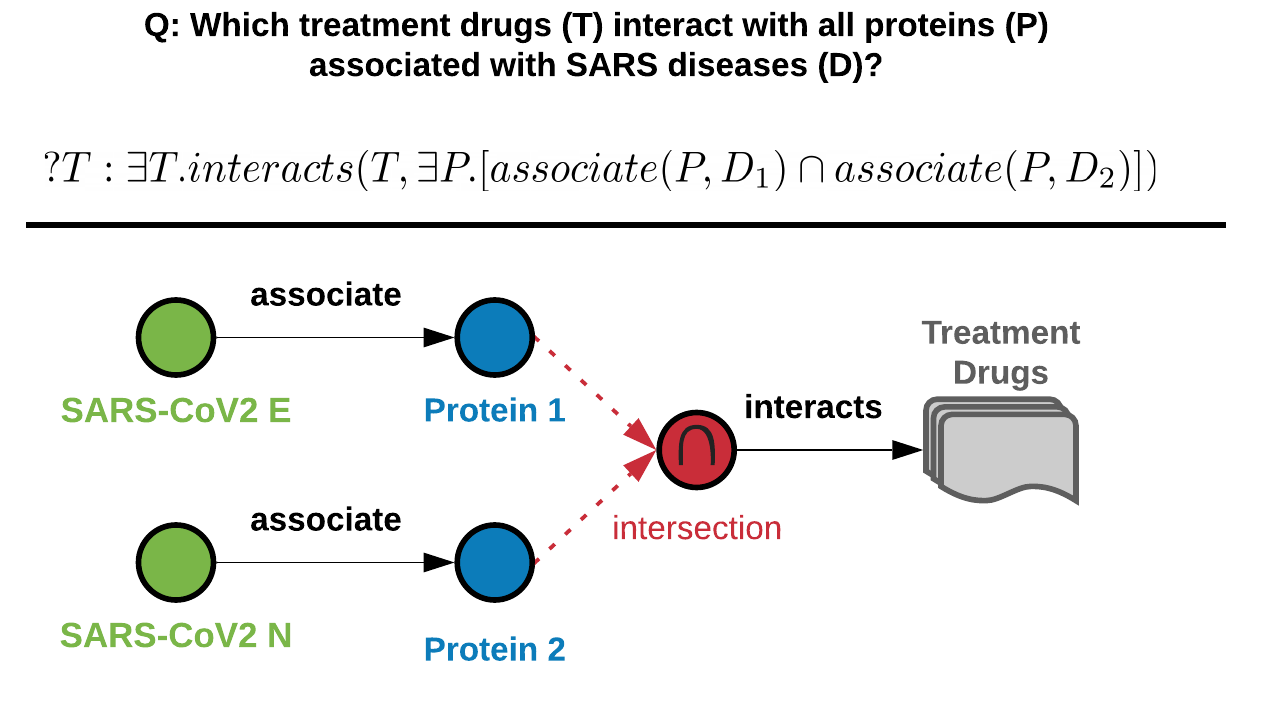}
     \vspace{-.8em}

         \caption{{Drug Repurposing (DRKG)}.}
         \label{fig:query_drkg_1}
     \end{subfigure}
     \begin{subfigure}[b]{0.47\linewidth}
         \centering
         \includegraphics[width=\linewidth]{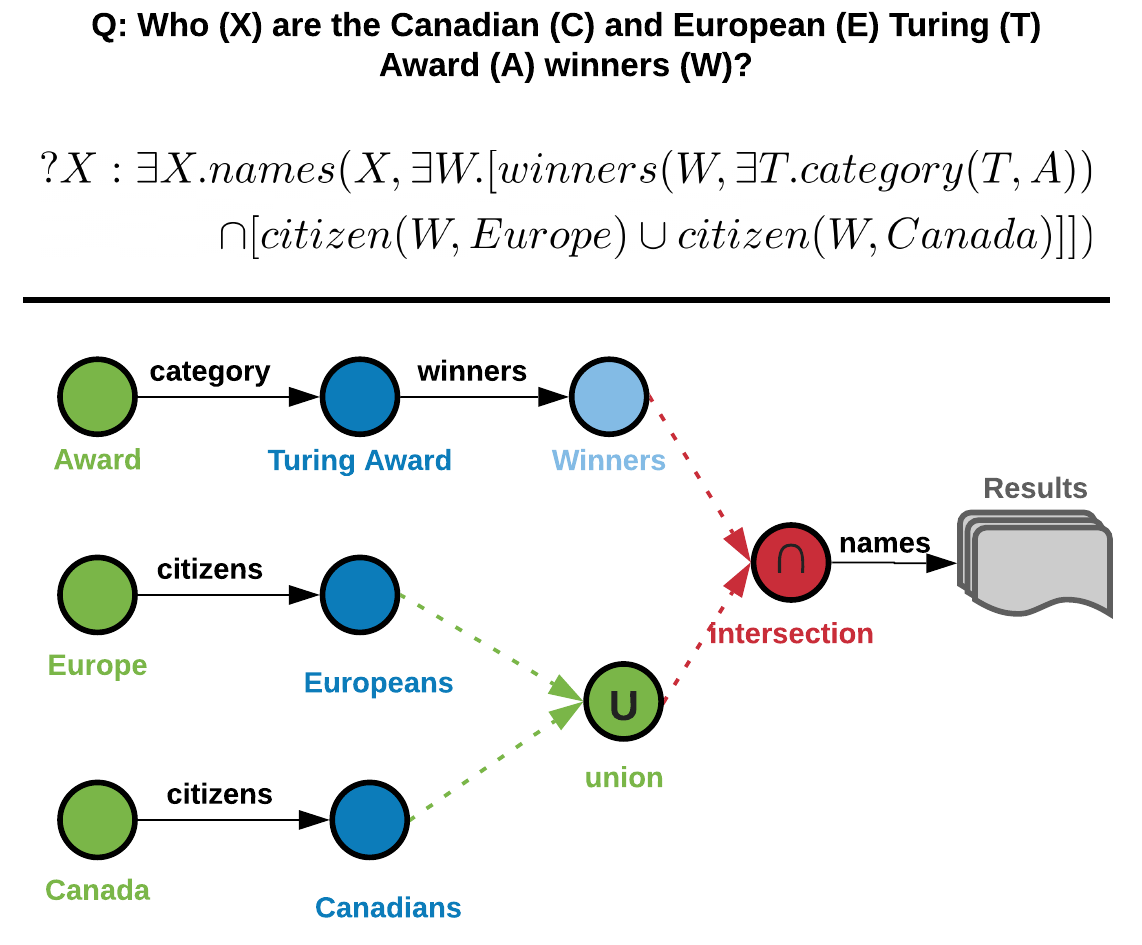}
     \vspace{-.8em}

         \caption{{Open-domain (FB15K)}.}
         \label{fig:query_fb15k}
     \end{subfigure}
    % \begin{subfigure}[b]{0.49\linewidth}
    %      \centering
    %      \includegraphics[width=\linewidth]{images/query_3.png}
    %           %%\vspace{-.8em}

    %      \caption{{Organized information (DBPedia).}}
    %      \label{fig:query_dbpedia}
    %  \end{subfigure}
     %%\vspace{-.4em}
        \caption{Sample FOE queries from different datasets that utilize existential quantification ($\exists$), intersection ($\cap$), and union ($\cup$) operations. The simple operations need to be chained together in an end-to-end {objective function} to retrieve relevant results for complex queries.}
        \label{fig:foe}
        \vspace{-1.2em}
\end{figure*}

Knowledge Graphs (KGs) are structured heterogeneous graphs where information is organized as triplets of entity pair and the relation between them. This organization provides a fluid schema with applications in several domains including e-commerce \cite{dong2020autoknow}, web ontologies \cite{bizer2009dbpedia,bonatti_et_al:DR:2019:10328}, and medical research \cite{LI2020101817,shi2017semantic}. Chain reasoning is a fundamental problem in KGs, which involves answering a chain of first-order existential (FOE) queries (translation, intersection, and union) using the KGs' relation paths. A myriad of queries can be answered using such logical formulation (some examples are given in Figure \ref{fig:foe}). Current approaches \cite{hamilton2018embedding,Ren*2020Query2box:, choudhary2020self} in the field rely on mapping the entities and relations onto a representational latent space such that the FOE queries can be reduced to mathematical operations in order to further retrieve the relevant answer entities.   

Euclidean vectors \cite{hamilton2018embedding,trouillon2016complex} provide a nice mechanism to encode the semantic position of the entities by leveraging their neighborhood relations. They utilize a fixed threshold over the vector to query for answer entities {(such as a k-nearest neighbor search)}. However, queries differ in their breadth. Certain queries would lead to a greater set of answers than others, e.g., query \texttt{Canadians} will result in a higher number of answers than query \texttt{Canadian Turing Award winners}. To capture this query behavior, spatial embeddings \cite{Ren*2020Query2box:,choudhary2020self,arakelyan2021complex,ren2020beta} {learn a border parameter that accounts for broadness of queries by controlling the volume of space enclosed by the query representations}. However, these spatial embeddings rely on more complex geometries such as boxes \cite{Ren*2020Query2box:} which do not have a {closed form solution} to the union operation, {e.g., the union of two boxes is not a box. Thus, further FOE operations cannot be applied to the union operation.} %to a union of boxes}. 
Additionally, their strict borders lead to some ambiguity in the border case scenarios and a non-smooth distance function, e.g., a point on the border will have a much smaller distance if it is considered to be inside the box than if it is considered to be outside. This challenge also applies to other geometric enclosures such as hyperboloids \cite{choudhary2020self}. %\nr{The same applies to other representations considered (circles, hyperboloids)? This paragraph seems like we're only saying what's wrong with Q2B. }

Another line of work includes the use of structured geometric regions \cite{erk2009representing,Ren*2020Query2box:} or {density functions} \cite{vilnis-etal-2018-probabilistic,vilnis2015word2gauss,ren2020beta,bojchevski2018deep} instead of vector points for representation learning. While these approaches utilize the representations for modeling individual entities and relations between them, we aim to provide a closed form solution to logical queries over KGs using the Gaussian density function which enables chaining the queries together. Another crucial difference in our work is in handling a stream of queries. Previous approaches rely on Disjunctive Normal Form (DNF) transformation which requires the entire query input. In our model, every operation is closed in the Gaussian space and, thus, operations of a large query can be handled individually and aggregated together for the final answers. %\nr{what do you mean "entire query input?"}
 %\nr{it will be nice to have an example of query stream, to clarify what exactly does it mean}
\begin{figure*}[htbp]
     \centering
     %%\vspace{-.8em}
     \begin{subfigure}[b]{0.32\linewidth}
         \centering
         \includegraphics[width=\linewidth]{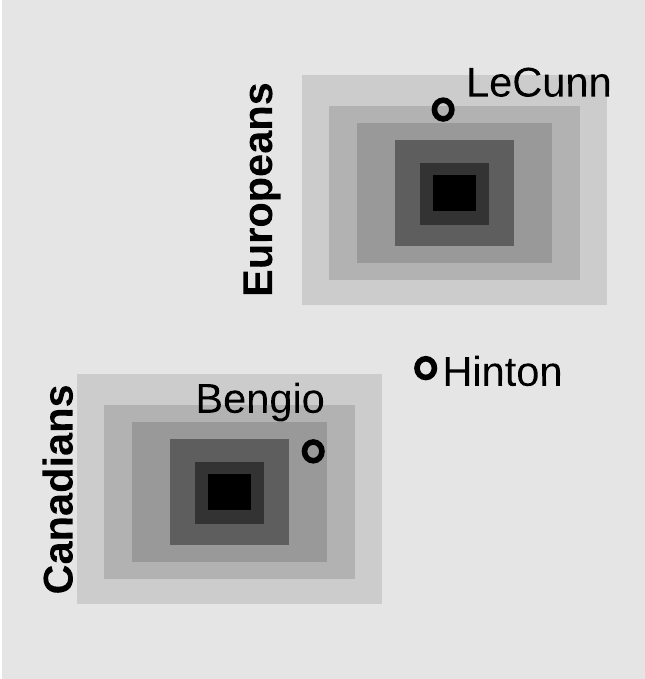}
         \caption{Union of box queries.}
         \label{fig:intro_box}
     \end{subfigure}
     \begin{subfigure}[b]{0.305\linewidth}
         \centering
         \includegraphics[width=\linewidth]{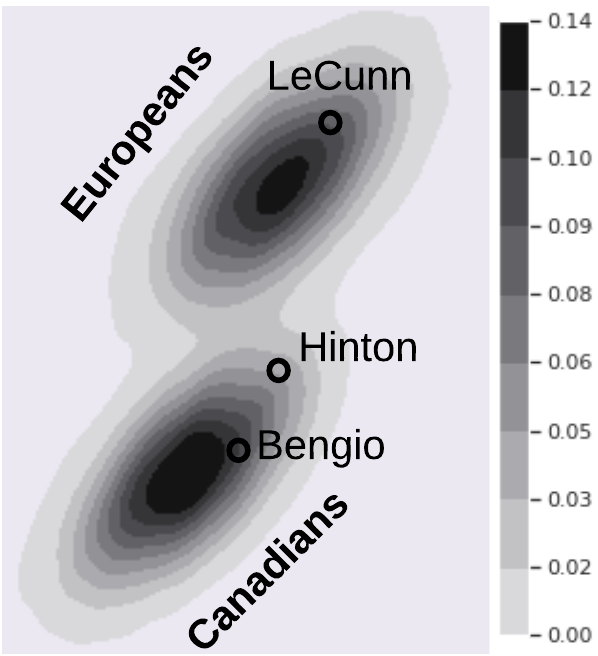}
         \caption{Gaussian mixture queries.}
         \label{fig:intro_Gaussian}
     \end{subfigure}
    \begin{subfigure}[b]{0.335\linewidth}
         \centering
         \includegraphics[width=\linewidth]{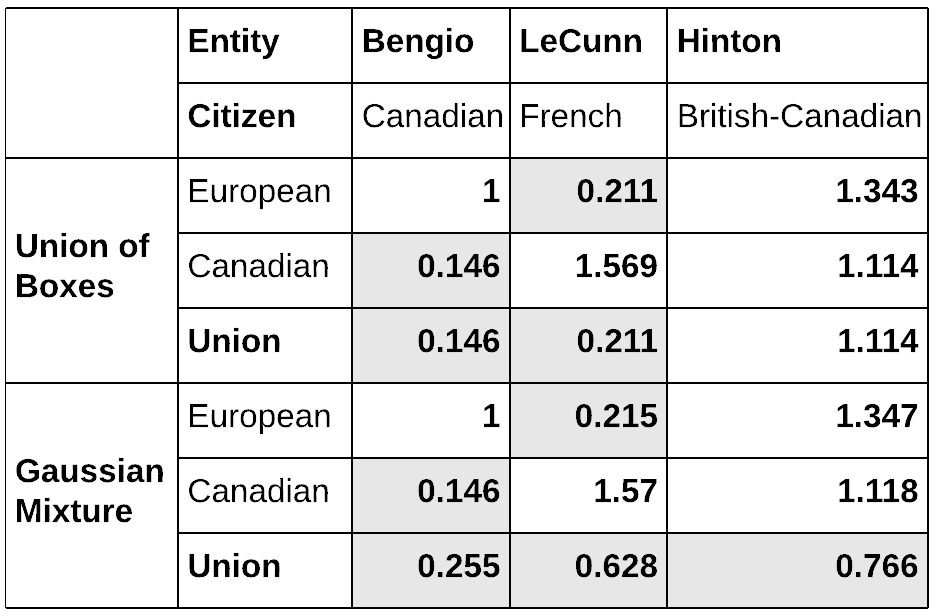}
         \caption{Distance of entities from query space. For comparability, distances are given relative to entity \texttt{Bengio}'s distance to \texttt{European} query.} %\nr{hard to read the numbers in darker boxes. Why shade them?}}
         \label{fig:intro_dist}
     \end{subfigure}
     %%\vspace{-.6em}
        \caption{Results of the query \texttt{Europeans}~$\cup$~\texttt{Canadians}. Entities in the darker areas have higher probability of being the answers than lighter areas. We can observe from (c) that the {non-smooth} borders of box geometry do not encompass the answer \texttt{Hinton}.}
        %%\vspace{-1em}
        \label{fig:intro_contrast}
\end{figure*}

To alleviate the drawbacks of operations not being closed under unions and border ambiguities, we propose {\fullmodel} (\model). {\model} models entities as {a mixture of Gaussian densities}. Gaussian densities have been previously used in natural language processing \cite{vilnis2015word2gauss} and graphs \cite{bojchevski2018deep} to enable more expressive parameterization of decision boundaries. In our case, we utilize a mixture of multivariate Gaussian densities due to their intuitive closed form solution for translation, intersection, and union operations. In addition, they can also enable the use of a smooth distance function; Mahalanobis distance \cite{mahalanobis}. Figure \ref{fig:intro_contrast} provides an example of such a case where the {non-smooth} boundaries of box query embeddings are not able to capture certain answers.
We utilize the mean ($\mu$) and co-variance ($\Sigma$) parameters of multivariate Gaussian densities to encode the semantic position and spatial query area of an entity, %\nr{do entities have query area or just queries? in HYPE the entities were points and the query was hyperboloid}
 respectively. The closed form solution for the operations allows us to solve complex queries by chaining them in a pipeline. {\model} does not need to rely on DNF transformations, since all the outputs are closed {in} the Gaussian space and complex queries can be consolidated in an end-to-end objective function, %\nr{this line should go somewhere in the next paragraph. You're mentioning PERM before introducing it (you do it in the next para)}
 e.g., in Figure \ref{fig:intro_Gaussian}, \texttt{Europeans}~$\cup$~ \texttt{Canadians} is a Gaussian mixture and the {single} objective is to minimize the distance between the mixture and entity \texttt{Hinton}, whereas in the case of boxes (shown in Figure \ref{fig:intro_box}), we have two independent objectives to minimize the distance from each box in the union query.
 %\nr{give an example here where Q2B fails and PERM succeeds}  
%Summarizing, our work advocates for Gaussian densities as a means for modeling KGs due to their following advantages:
Summarizing, the contributions of our work is as follows:
\begin{enumerate}[leftmargin=*]
    \item We develop \fullmodel~(\model), a method to reason over KGs using (mixture of) Gaussian densities. Gaussians are able to provide a closed form solution to intersection and union, and also a smooth distance function. This enables us to process a chain of complex logical queries in an end-to-end objective function. 
    \item {\model} is able to outperform the current state-of-the-art baselines on logical query reasoning over standard benchmark datasets. Additionally, it is also able to %use the SARS related diseases, to 
    provide better drug recommendations for COVID-19.
    \item {\model} is also interpretable since the Gaussian embeddings can be visualized after each query process to understand the complete query representation. 
\end{enumerate}

The rest of the paper is organized as follows: Section \ref{sec:related} presents the current work in the field. In section \ref{sec:operators}, we present {\model} and define its various operations. Section \ref{sec:chains} provides the formulation for building the reasoning chains for complex queries. We provide the experimental setup and results in section \ref{sec:exp}. We conclude our paper in section \ref{sec:conclusion} and present its broader impact in section \ref{sec:broader}. %\nr{if you run out of space you can remove this para}

\section{Related Work}
\label{sec:related}
The topic of multi-hop chain reasoning over KGs has gained a lot of attention in recent years \cite{bordes2013translating,nickel2011three,das-etal-2017-chains,hamilton2018embedding}. These approaches utilize vector spaces to model query representation and retrieve results using a fixed threshold. While such representations are efficient at encoding semantic information, the fixed thresholds that are typically used in these models do not allow for an expressive (adjustable) boundary and, thus, are not best suited for representing queries. Spatial embeddings \cite{Ren*2020Query2box:,choudhary2020self,sun2020faithful} enhance the simple vector representations by adding a learnable border parameter that controls the spatial area around a query representation. These methods have strict borders that rely on non-smooth distance function that creates ambiguity between border cases. On the other hand, in our model, the variance parameter of the query's Gaussian densities creates soft smoothly increasing borders in terms of the Mahalanobis distance. Additionally, the previous methods do not provide a closed form solution for unions which we solve using Gaussian mixture models.

Density-based embeddings have seen a recent surge of interest in various domains. Word2Gauss \cite{vilnis2015word2gauss} provides a method of learning Gaussian densities for words from their distributional semantic information. In addition, the authors further apply this work to knowledge graphs \cite{vilnis-etal-2018-probabilistic}. Another approach \cite{bojchevski2018deep} aims to learn Gaussian graph representations from their network connections. These methods are, however, focused on learning semantic information and do not easily extend to logical queries over knowledge graphs. {\model} primarily focuses on learning spatial Gaussian densities for queries, while also capturing the semantic information. To achieve this, we derive closed form solutions to FOE queries.
%In that sense, our work is also related to classic multivariate Gaussian distributions.} %\nr{not sure if these 2 citations are needed in "related work"} \cite{tong2012multivariate,petersen2008matrix}.}
%\nr{An alternative can be to beef up the intro with these density-dependent related work, and remove this section. If you're able to cover all the related work and it's drawbacks compared to PERM then this section can be redundant. }

\section{{\fullmodel} for Logical Operators}
\label{sec:operators}
Knowledge Graphs (KG) $\mathcal{G}:E\times R$ are heterogeneous graphs that store entities ($E$) and relations ($R$). Each relation $r \in R$ is a Boolean function $r:E\times E \rightarrow \{True,False\}$ that indicates if the relation $r$ exists between a pair of entities. Without loss of generality, KGs can also be organized as a set of triples $\langle e_1,r,e_2\rangle \subseteq \mathcal{G}$, defined by the Boolean relation function $r(e_1,e_2)$. In this work, we focus on the following three FOE operations: translation (t), intersection ($\cap$), and union ($\cup$). The operations are defined as below:
%{\color{red}SK: should we explain how to read these definitions or give an example of what Qt, qt, Vt are?} \nr{+1 to sumeet. Or add a line referring the users to some other paper where they've defined this with examples}
\begin{align}
    q_t[Q_t] &\delequal?V_t: \{v_1,v_2,...,v_k\} \subseteq E ~ \exists ~ a_1\\
    q_\cap[Q_\cap] &\delequal?V_\cap: \{v_1,v_2,...,v_k\}  \subseteq E ~ \exists ~ a_1 \cap a_2 \cap ... \cap a_i\\
    q_\cup[Q_\cup] &\delequal?V_\cup: \{v_1,v_2,...,v_k\} \subseteq E ~ \exists ~ a_1 \cup a_2 \cup ... \cup a_i\\
    \text{where }Q_t & = (e_1,r_1);~ Q_\cap, Q_\cup = \{(e_1,r_1),(e_2,r_2),..(e_i,r_i)\} \nonumber 
    \text{~~and~ }a_i = r_i(e_i,v_a) \nonumber
\end{align}
%\nr{is the above notation standard? with LHS being query and RHS being result? mathematically this means $q = V$ no? equations (1), (2) etc in 10 are more intuitive i think. } 
where $q_t$, $q_\cap$, and $q_\cup$ are the translation, intersection, and union queries, respectively; and $V_t$, $V_\cap$, and $V_\cup$ are the corresponding results \cite{arakelyan2021complex}.
As we notice above, each entity has a dual nature; one as being part of a query and another as a candidate answer to a query.
In {\model}, we model the query space of an entity $e_i \in E$ as a multivariate Gaussian density function; $e_i = \mathcal{N}(\mu_i,\Sigma_i)$, where the learnable parameters $\mu_i$ (mean) and $\Sigma_i$ (covariance) indicate the semantic position and the surrounding query density of the entity, respectively. As a candidate, we only consider the $\mu_i$ and ignore the $\Sigma_i$ of the entity. We define the distance of a candidate entity %\nr{should these be $v_i \sim \mathcal{N}()$ ? i guess $=$ is fine since you're not sampling } 
$v_i=\mathcal{N}(\mu_i,\Sigma_i)$ from a query Gaussian $e_j = \mathcal{N}(\mu_j,\Sigma_j)$ using the Mahalanobis distance \cite{mahalanobis} given by:%{\color{red}SK: do you inted to define the distance here?}
\begin{align}
    d_\mathcal{N}(v_i,e_j) = (\mu_j-\mu_i)^T\Sigma_j^{-1}(\mu_j-\mu_i)\label{eq:mahalanobis}
\end{align}

\begin{figure*}[htbp]
%%\vspace{-.8em}
        
     \centering
     \begin{subfigure}[b]{0.32\linewidth}
         \centering
         \includegraphics[width=\linewidth]{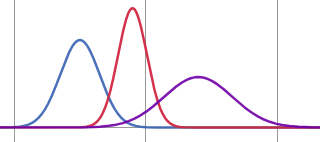}
         %%\vspace{-.8em}
        
         \caption{Translation ($q_t$)}
         \label{fig:translation}
     \end{subfigure}
     \begin{subfigure}[b]{0.32\linewidth}
         \centering
         \includegraphics[width=\linewidth]{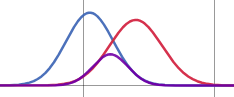}
        %\vspace{-.8em}
        
         \caption{Intersection ($q_\cap$)}
         \label{fig:intersection}
     \end{subfigure}
    \begin{subfigure}[b]{0.32\linewidth}
         \centering
         \includegraphics[width=\linewidth]{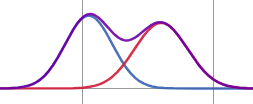}
        %\vspace{-.8em}
        
         \caption{Union ($q_\cup$)}
         \label{fig:union}
     \end{subfigure}
     \centering
     \begin{subfigure}[b]{0.32\linewidth}
         \centering
         \includegraphics[width=\linewidth]{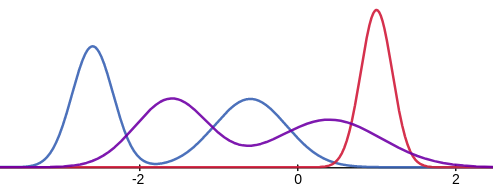}
        %\vspace{-.8em}
        
         \caption{Chain Translation ($c_t$)}
         \label{fig:chain_translation}
     \end{subfigure}
     \begin{subfigure}[b]{0.32\linewidth}
         \centering
         \includegraphics[width=\linewidth]{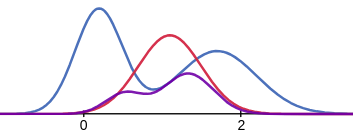}
         \caption{Chain Intersection ($c_\cap$)}
         \label{fig:chain_intersection}
     \end{subfigure}
    \begin{subfigure}[b]{0.32\linewidth}
         \centering
         \includegraphics[width=\linewidth]{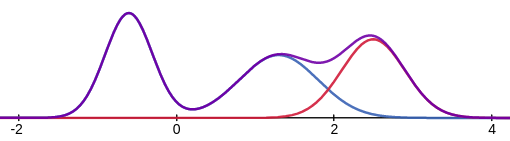}
         \caption{Chain Union ($c_\cup$)}
         \label{fig:chain_union}
        \end{subfigure}
        %\vspace{-.8em}
        \caption{The logical single (top row) and chain operations (bottom row) of translation, intersection, and union in the Gaussian space. The operations are closed and will result in either a Gaussian density or a Gaussian mixture. The input operands are given in {\color{Gaussian_blue}{blue}} and {\color{Gaussian_red}{red}} and the resultant Gaussian density{/mixture} is depicted in {\color{Gaussian_purple}{purple}}. For simplicity, the example is given for a univariate Gaussian model, but in our work, we use multivariate Gaussian densities.}
        %\vspace{-1.2em}
        \label{fig:operations}
\end{figure*}

Additionally, we need to define the FOE operations for the proposed {\fullmodel}. A visual interpretation of the operations; translation, intersection, and union is shown in Figure \ref{fig:operations}. The operations are defined as follows:

\textbf{Translation (t).} Each entity $e \in E$ and $r \in R$ are encoded as $\mathcal{N}(\mu_e,\Sigma_e)$ and $\mathcal{N}(\mu_r,\Sigma_r)$, respectively.  We define the translation query representation of an entity $e$ with relation $r$ as $q_t$ and the distance of resultant entity $v_t \in V_t$ from the query as $d^q_t$ given by: %{\color{red}SK: $o_t$ is defined but not used}
\begin{align}
    q_t &= \mathcal{N}(\mu_e+\mu_r,(\Sigma_e^{-1}+\Sigma_r^{-1})^{-1});\quad d^q_t = d_\mathcal{N}(v_t,q_t)
    \label{eq:translation}
\end{align}
\textbf{Intersection ($\cap$).} Intuitively, the intersection of two Gaussian densities implies a random variable that belongs to both the densities. Given that the entity densities are independent of each other, we define the intersection of two entity density functions $e_1, e_2$ as $q_\cap$ and distance of resultant entity $v_\cap \in V_\cap$ from the query as $d^q_\cap$ given by:
\begin{align}
    q_\cap &= \mathcal{N}(\mu_{e_1}, \Sigma_{e_1})\mathcal{N}(\mu_{e_2}, \Sigma_{e_2}) =
    \mathcal{N}(\mu_3, \Sigma_3);\quad d^q_\cap = d_\mathcal{N}(v_\cap,q_\cap) \label{eq:q_intersection}\\
    \text{where, } &\Sigma_3^{-1} = \Sigma_1^{-1} + \Sigma_2^{-1} \nonumber\\
    \text{and }&\mu_3 = \Sigma_3(\Sigma_2^{-1}\mu_1+\Sigma_1^{-1}\mu_2)%\text{ (or) }
    \implies 
    \Sigma_3^{-1}\mu_3 = \Sigma_2^{-1}\mu_1+\Sigma_1^{-1}\mu_2 \nonumber 
\end{align}

We provide a brief sketch of the proof that the intersection of Gaussian density functions is a closed operation. A complete proof is provided in Appendix \ref{app:gaussian_product}.
%{\textit{Derivation that the intersection of Gaussian density functions is a closed operation\footnote{Here, we present a proof sketch. Complete proof is provided in Appendix \ref{app:gaussian_product}}.}} 
Let us consider two Gaussian PDFs $P(\theta_1)=\mathcal{N}(\mu_1,\Sigma_1)$ and $P(\theta_2)=\mathcal{N}(\mu_2,\Sigma_2)$. Their intersection implies a random variable that is distributed as the product, $P(\theta_1)P(\theta_2)$ %\nr{im not sure of using the term "part of 2 distributions". Should we just say it's an RV distributed as the product? what do you guys think?}
The intersection $P(\theta)=\mathcal{N}(\mu_3,\Sigma_3)$ is derived as follows:
\begin{align}
    P(\theta)&=P(\theta_1).P(\theta_2)\nonumber\\
    \log(P(\theta)) &= (x-\mu_1)^T\Sigma_1^{-1}(x-\mu_1) + (x-\mu_2)^T\Sigma_2^{-1}(x-\mu_2)\nonumber\\
    (x-\mu_3)^T\Sigma_3^{-1}(x-\mu_3) &= (x-\mu_1)^T\Sigma_1^{-1}(x-\mu_1) + (x-\mu_2)^T\Sigma_2^{-1}(x-\mu_2)\nonumber \\
    \text{Comparing coefficients;}&\quad
    \Sigma_3^{-1} = \Sigma_1^{-1} + \Sigma_2^{-1};\quad\mu_3 = \Sigma_3(\Sigma_2^{-1}\mu_1+\Sigma_1^{-1}\mu_2)\nonumber
\end{align}

\textbf{Union ($\cup$).} We model the union of multiple entities using Gaussian mixtures. The union of entity density functions given by $e_1, e_2, e_3, ..., e_n$ is defined as $q_\cup$ and the distance of resultant entity $v_\cup \in V_\cup$ from the query as $d^q_\cup$ given by:%{\color{red}SK:what is $\phi$ in the below equations?} \nr{can you also derive the union of Gaussian mixtures to be another mixture?}
\begin{align}
    q_\cup &= \sum_{i=1}^n\phi_i\mathcal{N}(\mu_{e_i},\Sigma_{e_i});\quad
    d^q_\cup = \sum_{i=1}^n\phi_id_\mathcal{N}(v_\cup,\mathcal{N}(\mu_{e_i},\Sigma_{e_i})) \label{eq:q_union}\\
    \text{where, }\phi_i &= \frac{exp\left(\mathcal{N}(\mu_{e_i},\Sigma_{e_i})\right)}{\sum_{j=1}^nexp\left(\mathcal{N}(\mu_{e_j},\Sigma_{e_j})\right)} \nonumber
\end{align}
$\phi_i \in \Phi$ are the weights for each Gaussian density in the Gaussian mixture, calculated using the self-attention mechanism over the parameters of the Gaussians in the mixture, i.e., $\mu_{e_i},\Sigma_{e_i} \forall i:1\rightarrow n$. 

\section{Chain Reasoning over Knowledge Graphs}
\label{sec:chains}
{We consider the Gaussian density function {(embedding of a single entity)} as a special case of Gaussian mixture with a single component.
This ensures that all }the operations defined in Section \ref{sec:operators} are closed under the Gaussian space with an output that is either a single (for translations and intersections) or multi-component Gaussian mixture (for unions). Hence, for chaining the queries, we need to define the logical operators with a Gaussian density and a Gaussian mixture input. In this section, we define the different operators (depicted in Figure \ref{fig:operations}), in the case of a Gaussian mixture input.

\textbf{Chain Translation.} Let us assume that the input query embedding is an $n$-component mixture $p=\sum_{i=1}^n\mathcal{N}(\mu_i,\Sigma_i)$ and we need to translate it with relation $r = \mathcal{N}(\mu_r,\Sigma_r)$. %\nr{careful about what subscripts you use. you've used $i$ in 2 places here, and also to denote entities above}. 
Intuitively, we would like to translate all the Gaussians in the mixture with the relation. Hence, we {model} this translation as $c_t$ and the distance from entities $v_t \in V_t$ as $d^c_t$ given by: %\nr{by calling these "definitions", it seems like you're coming up with these things. They should be derivations no? How did you get eqn (8)?. Same for chain intersections and unions}
\begin{align}
    c_t &= \sum_{i=1}^n\phi_i\mathcal{N}(\mu_i+\mu_r,(\Sigma_i^{-1}+\Sigma_r^{-1})^{-1})\\
    d^c_t &= \sum_{i=1}^n\phi_id_\mathcal{N}(v_t,\mathcal{N}(\mu_i+\mu_r,(\Sigma_i^{-1}+\Sigma_r^{-1})^{-1}) 
\end{align}

\textbf{Chain Intersection.} A Gaussian mixture is a union over individual densities. Based on the distributive law of sets, an intersection over a Gaussian mixture $p=\sum_{i=1}^n\mathcal{N}(\mu_i,\Sigma_i)$ and entity $e=\mathcal{N}(\mu_e,\Sigma_e)$ implies the union of the intersection between the entity and each Gaussian density in the mixture. Hence, we {derive} this intersection as $c_\cap$ and the distance from entities $v_\cap \in V_\cap$ as $d^c_\cap$:
\begin{align}
    \nonumber c_\cap &= \cup_{i=1}^n\mathcal{N}(\mu_{e}, \Sigma_{e})\mathcal{N}(\mu_i, \Sigma_i) = \sum_{i=1}^n\phi_i\mathcal{N}(\mu_{e}, \Sigma_{e})\mathcal{N}(\mu_i, \Sigma_i) \\
    \implies c_\cap&= \sum_{i=1}^n\phi_i\mathcal{N}(\mu_{e\cap i}, \Sigma_{e\cap i}) \label{eq:c_intersection}\\
    \text{where, }&\Sigma_{e\cap i}^{-1} = \Sigma_{e}^{-1} + \Sigma_{i}^{-1}~\text{ and }~\mu_{e\cap i} = \Sigma_{e\cap i}(\Sigma_i^{-1}\mu_e+\Sigma_e^{-1}\mu_i) \nonumber\\
    d^c_\cap &= \sum_{i=1}^n\phi_id_\mathcal{N}(v_\cap,\mathcal{N}(\mu_{e\cap i}, \Sigma_{e\cap i}))
\end{align}

\textbf{Chain Union.} The union of an entity $e = \mathcal{N}(\mu_e,\Sigma_e)$ with a Gaussian mixture $\sum_{i=1}^n\phi_i\mathcal{N}(\mu_i,\Sigma_i)$ is the addition of the entity to the mixture. Hence, the  union $c_\cup$ and the distance from entities $v_\cup \in V_\cup$ $d^c_\cup$ can be defined as follows:
\begin{align}
    c_\cup &= \sum_{i=1}^n\phi_i\mathcal{N}(\mu_i, \Sigma_i)+\phi_e\mathcal{N}(\mu_e, \Sigma_e)\\
    d^c_\cup &= \sum_{i=1}^n\phi_id_\mathcal{N}(v_\cup,\mathcal{N}(\mu_{i},\Sigma_{i}))+\phi_ed_\mathcal{N}(v_\cup,\mathcal{N}(\mu_{e},\Sigma_{e}))
\end{align}

\textbf{Implementation Details.} To calculate the weights ($\phi_i \in \Phi$) of the Gaussian mixtures, we use the popular self-attention mechanism \cite{vaswani2017attention}. The gradient descent over Mahalanobis distance (Eq. \ref{eq:mahalanobis}) and derivation for the product of Gaussians (Eq. \ref{eq:q_intersection}) are given by \cite{law2016closed} and Appendix \ref{app:gaussian_product}, respectively. {Another important note is that we do not need to compute $\Sigma$ for the operations, but rather we only need to compute the $\Sigma^{-1}$.} Also, storing the complete $\Sigma^{-1}$ requires quadratic memory, i.e., a Gaussian density of $d$ variables requires $d \times d$ parameters for $\Sigma$. So, we only store a decomposed matrix $L$ of $\Sigma^{-1}:\Sigma^{-1}=LL^T$. Thus, for a Gaussian density of $d$ variables our memory requirement is $d\times(r+1)$ parameters ($d$ for $\mu$ and $d\times r$ for $\Sigma^{-1}$). %dimensions of L are dx1, we didn't need any bigger matrix\nr{what is the dimension of $L$? how is it only $d$? shouldnt it be $dk$ params for each $L$ where $k$ is the rank?}. 
For computing the $\mu_3$ for intersection, in Eq. (\ref{eq:q_intersection}), we use a linear solver (\texttt{torch.solve}) %\nr{any specific solver? cite?} 
for faster computation. All our models are implemented in Pytorch \cite{pytorch} and run on four Quadro RTX 8000. \footnote{Implementation code: \url{https://github.com/Akirato/PERM-GaussianKG}}

%\nr{i recommend moving some derivations to the main section and move them out when we are actually short on space. doesn't look like space is an issue}

\section{Experiments}
\label{sec:exp}
This section describes the experimental setup used to analyze the performance of {\model} on various tasks with a focus on the following research questions:
\begin{enumerate}[leftmargin=*]
    \item Does {\model}'s query representations perform better than the state-of-the-art baselines on the task of logical reasoning over standard benchmark knowledge graphs?
    \item What is the role of individual components in {\model}'s overall performance gain?
    \item Is {\model} able to recommend better therapeutic drugs for COVID-19 from drug re-purposing graph data {compared to the current baselines}?
    %\nr{better with respect to?}
    \item Are we able to visualize the {operations on} {\model}'s query representations in the latent space?
\end{enumerate}

\subsection{Datasets and Baselines}
We utilize the following standard benchmark datasets to compare {\model}'s performance on the task of reasoning over KGs:
\begin{itemize}[leftmargin=*]
    \item \textbf{FB15K-237} \cite{toutanova2015representing} is comprised of the 149,689 relation triples and textual mentions of Freebase entity pairs. All the simply invertible relations are removed. %The total number of entity and relation types are 14,505 and 237, respectively. 
    \item \textbf{NELL995} \cite{carlson2010toward} consists of 107,982 triples obtained from the $995^{th}$ iteration of the Never-Ending Language Learning (NELL) system. %The KG is comprised of 63,361 entities and 200 relations.
    \item \textbf{DBPedia}\footnote{\url{https://www.kaggle.com/danofer/dbpedia-classes}} is a subset of the Wikipedia snapshot that consists of a multi-level hierarchical taxonomy over 240,942 articles. %The taxonomy consists of 34,575 entities and 3 relation types.
    \item  \textbf{DRKG} \cite{drkg2020} (Drug Re-purposing Knowledge Graph) is used to evaluate the performance of our model on both the logical reasoning and drug recommendation tasks. %It consists of 5,874,261 relation triples with 97,238 entities and 107 edges. \nr{these details are already in Table 1, so you can save space by not mentioning num nodes, num edges etc in these bullet points. Can just have the 4 datasets in one line and point to Table 1 for the statistics.}
\end{itemize}
\begin{table}[htbp]
		\caption{Dataset statistics including the number of unique entities, relations, and edges, along with the splits of dataset triples used in the experiments.}%\protect\footnotemark}
	\begin{tabular}{|l|ccc|ccc|}
		\hline
		\textbf{Dataset}&\textbf{\# Entities}&\textbf{\# Relations}&\textbf{\# Edges}&\textbf{\# Training}&\textbf{\# Validation}&\textbf{\# Test}\\
		\hline
		FB15k-237&14,505&237&310,079&272,115&17,526&20,438\\
		NELL995&63,361&200&142,804&114,213&14,324&14,267\\
		DBPedia&34,575&3&240,942&168,659&24,095&48,188\\
		DRKG&97,238 &107&5,874,271&4,111,989&587,428&1,174,854\\
		\hline
	\end{tabular}
	\label{tab:dataset_stats}
\end{table} 
More detailed statistics of these datasets are provided in Table \ref{tab:dataset_stats}. For our experiments, we select the following baselines based on (i) their performance on the logical reasoning task and (ii) their ability to extend to all FOE query combinations.
\begin{itemize}[leftmargin=*]
    \item \textbf{Graph Query Embedding (GQE)}  \cite{hamilton2018embedding} embeds entities and relations as a vector and utilizes TransE \cite{bordes2013translating} to learn the query embeddings. The distance of the answer entities is calculated using L1-norm.
%     \item \textbf{Knowledge Graph Attention Network} (KGAT) \cite{wang2019kgat} embeds
% entities and relations as vectors and utilizes attention networks over entities and relations
% with a TransR loss \cite{lin2015learning}.
\item \textbf{Query2Box (Q2B)} \cite{Ren*2020Query2box:} embeds entities and relations as axis aligned hyper-rectangles or boxes and utilize FOE queries to learn query representations. The distance of answer entities is given by a weighted combination of the answer's distance from the center and the border of the query box.
\item \textbf{Beta Query Embedding (BQE)} \cite{ren2020beta} utilizes beta distribution to learn query representations from FOE queries with a novel addition of negation queries. The distance is calculated as the dimension-wise KL divergence between the answer entity and the query beta embedding.
\item \textbf{Complex Query Decomposition (CQD)} \cite{arakelyan2021complex} answers complex queries by reducing them to simpler sub-queries and aggregating the resultant scores with t-norms. 
\end{itemize}
Some of the other baselines \cite{yang2014embedding,nickel2011three} focus solely on the multi-hop problem. They could not be intuitively extended to handle all FOE queries, and hence, we did not include them in our study.
\begin{table*}[htbp]
	\centering
	\caption{Performance comparison of {\model} against the baselines to study the efficacy of the query representations. The columns present the different query structures and the overall average performance. The last row presents the Average Relative Improvement (\%) of {\model} compared to CQD over all datasets across different query types. Best results for each dataset are shown in bold. The MRR results for experiments are given in Appendix \ref{app:mrr}.}
	\vspace{-.8em}
	\begin{tabular}{|ll|ccc|ccc|ccc|c|}
		\hline
		\textbf{}&&\multicolumn{10}{c|}{\textbf{HITS@3}}
		\\\hline
		\textbf{Dataset}&\textbf{Model}&\textbf{1t}&\textbf{2t}&\textbf{3t}&\boldmath$2\cap$&\boldmath$3\cap$&\boldmath$2\cup$&\boldmath$\cap t$&\boldmath$t\cap$&\boldmath$\cup t$&\textbf{Avg}\\\hline
FB15k-237&GQE&.404&.214&.147&.262&.390&.164&.087&.162&.155&.221\\
&BQE&.455&.122&.102&.232&.459&.141&.224&.124&.101&.218\\
&Q2B&.467&.240&.186&.324&.453&.239&.050&.108&.193&.251\\
&CQD&.512&\textbf{.288}&\textbf{.221}&.352&.457&.284&\textbf{.129}&\textbf{.249}&.121&.290\\
&{\model}&\textbf{.520}&.286&.216&\textbf{.361}&\textbf{.490}&\textbf{.305}&.128&.212&\textbf{.239}&\textbf{.306}\\\hline
NELL995&GQE&.417&.231&.203&.318&.454&.200&.081&.188&.139&.248\\
&BQE&\textbf{.711}&.156&.132&\textbf{.438}&.540&.153&\textbf{.250}&.160&.091&.292\\
&Q2B&.555&.266&.233&.343&.480&.369&.132&.212&.163&.306\\
&CQD&.667&\textbf{.350}&\textbf{.288}&.410&\textbf{.529}&\textbf{.531}&.171&\textbf{.277}&.156&\textbf{.375}\\
&{\model}&.581&.286&.243&.352&.508&.460&.143&.195&\textbf{.200}&.328\\\hline
DBPedia&GQE&.673&.006\footnotemark&N.A.&.873&.879&.402&.160&.668&0.00&.458\\
&BQE&.881&\textbf{.007}\footnotemark[\value{footnote}]&N.A.&\textbf{1.00}&\textbf{1.00}&.384&\textbf{.435}&.590&0.00&.565\\
&Q2B&.832&\textbf{.007}\footnotemark[\value{footnote}]&N.A.&\textbf{1.00}&\textbf{1.00}&.649&.224&.856&0.00&.571\\
&CQD&.870&\textbf{.007}\footnotemark[\value{footnote}]&N.A.&\textbf{1.00}&\textbf{1.00}&.673&.218&.787&0.00&.569\\
&{\model}&\textbf{.950}&\textbf{.007}\footnotemark[\value{footnote}]&N.A.&\textbf{1.00}&\textbf{1.00}&\textbf{.782}&.232&\textbf{.952}&0.00&\textbf{.615}\\\hline
DRKG&GQE&.420&.218&.153&.270&.409&.181&.101&.186&.174&.235\\
&BQE&.554&.141&.123&.347&.512&.185&\textbf{.281}&.173&.124&.271\\
&Q2B&.499&.263&.199&.337&.489&.284&.068&.134&.235&.279\\
&CQD&.554&\textbf{.323}&\textbf{.238}&.369&.495&.341&.184&\textbf{.310}&.150&.329\\
&{\model}&\textbf{.565}&.322&.236&\textbf{.387}&\textbf{.540}&\textbf{.376}&.190&.273&\textbf{.297}&\textbf{.354}\\\hline
\hline
\multicolumn{2}{|c|}{\textbf{{\model} vs Q2B (\%)}}&10.9&12.3&13.0&7.20&6.10&26.3&84.2&50.8&24.3&15.9\\
\multicolumn{2}{|c|}{\textbf{{\model} vs CQD  (\%)}}&3.80&-0.9&-2.4&2.00&5.80&9.50&1.80&-5.5&93.0&6.2\\\hline	
\end{tabular}
	\label{tab:efficiency}
%\vspace{-1em}
\end{table*}
\footnotetext{\label{note1}DBPedia has an extremely large number of resultant grand-children  leaves ($\approx 10^3$ per grand-parent) for the 2t task and, thus, we notice poor performance on 2t task across all the evaluation models.}
\subsection{(RQ1) Reasoning over KGs}
To evaluate the efficacy of {\model}'s query representations, we  compare it against the baselines on different FOE query types; (i) Single Operator: 1$t$, 2$t$, 3$t$, 2$\cap$, 3$\cap$, 2$\cup$ and (ii) Compound Queries: $\cap t$, $t \cap$, $\cup t$. We follow the standard evaluation protocol \cite{Ren*2020Query2box:, ren2020beta,choudhary2020self} and utilize the three splits of a KG for training $\mathcal{G}_{train}$, validation $\mathcal{G}_{valid}$, and evaluation $\mathcal{G}_{test}$ (details in Table \ref{tab:dataset_stats}). The models are trained on $\mathcal{G}_{train}$ with validation on $\mathcal{G}_{valid}$. The final evaluation metrics for comparison are calculated on $\mathcal{G}_{test}$. For the baselines, we calculate the relevance of the answer entities to the queries based on the distance measures proposed in their respective papers. In {\model}, the distance of the answer entity from the query Gaussian density is computed according to the measures discussed in Sections \ref{sec:operators} and \ref{sec:chains}. We use the evaluation metrics of HITS@K and MRR to compare the ranked set of results obtained from different models. Given the ground truth $\hat{E}$ and model outputs $\{e_1,e_2,...,e_n\} \in E$, the metrics are calculated as follows: %\nr{can move these definitions to appendix if you need space. }
\begin{align}
    \text{HITS@K}&= \frac{1}{K}\sum_{k=1}^{K}f(e_k);~f(e_k)=\begin{cases}1, \text{if }e_k \in \hat{E}\\
    0,\text{ else}\nonumber
    \end{cases}\\
    \text{MRR}&= \frac{1}{n}\sum_{i=1}^{n}\frac{1}{f(e_i)};~f(e_i)=\begin{cases}i,\text{if }e_i \in \hat{E}\\
    \infty,\text{ else}\nonumber
    \end{cases}
\end{align}

From the results provided in Table \ref{tab:efficiency}, we observe that {\model}, is able to outperform all the current state-of-the-art approaches, on an average across all FOE queries by $6.2\%$. Specifically, we see a consistent improvement for union queries; $9.5\%$ and $93\%$ in the case of $2\cup$ and $\cup t$, respectively. Comparing the models based on only geometries, we notice the clear efficacy of {\model} query representations with an average improvement of $37.9\%$, $15.9\%$, and $37.3\%$ over vectors (GQE), boxes (Q2B), and beta distribution (BQE), respectively. Given these improvements and the ability to handle compound queries in an end-to-end manner, we conclude that Gaussian distributions are better at learning query representations for FOE reasoning over KGs. Additionally, we provide {\model}'s results on sample queries from different datasets in Table \ref{tab:qualitative}.

\begin{table}[htbp]
    \centering
    \caption{Qualitative results of {\model} on samples from different datasets. Results given in \textcolor{ao_english}{green} and \textcolor{red}{red} indicate a correct and incorrect prediction, respectively.}
    \begin{tabular}{|p{20em}|p{18em}|}
    \hline
     \textbf{Query} &\textbf{Results} \\
    \hline
    Who are European and Canadian Turing awards winners?&\textcolor{ao_english}{Jeffrey Hinton}, \textcolor{ao_english}{Yoshua Bengio}, \textcolor{red}{Andrew Yao}\\
    Which Actors and Football Players also became Governors?&\textcolor{ao_english}{Arnold Schwarzenegger}, \textcolor{ao_english}{Heath Shuler}, \textcolor{red}{Frank White}\\
    Which treatment drugs interact with all proteins associated with SARS diseases?&\textcolor{ao_english}{Ribavirin}, \textcolor{ao_english}{Dexamethasone}, \textcolor{ao_english}{Hydroxychloroquine}\\
    \hline
    \end{tabular}
    \vspace{-1em}
    \label{tab:qualitative}
\end{table} 
\subsection{(RQ2) Ablation Study}
In this section, we evaluate the need for different components and their effects on the overall performance of our model. First, we look at the contribution of utilizing different types of queries to the performance of our model. For this, we train our model on different subsets of queries; (i) only 1$t$ queries, (ii) only translation (1$t$,2$t$,3$t$) queries and (iii) only single operator queries (1$t$,2$t$,3$t$,$2\cap$,$3\cap$,$2\cup$). Furthermore, we look at the need for attentive aggregation in the case of union of Gaussian mixtures. We test other methods of aggregation; (i) vanilla averaging and (ii) MLP \cite{murtagh1991multilayer}.

\begin{table*}[htbp]
	\centering
	\caption{Ablation study results. Performance comparison of {\model} (final) against different variants of our model. \textit{1t}, \textit{translation} and \textit{single} utilize the 1-hop queries, all translation queries and all single operator queries, respectively. The \textit{average} and \textit{MLP} variants utilize vanilla averaging and MLP for aggregation in union queries. {The metrics reported here are an average over all the datasets.} %\nr{why don't the final numbers match that in table 1? It's an average over all datasets. I had first planned on putting only FB15K-237. But, I was able to run the mode for all, so I put the average} 
	Finer evaluation with results for each dataset is given in Appendix \ref{app:ablation}. Best results are given in bold.}
	\vspace{-.8em}
	\begin{tabular}{|l|ccc|ccc|ccc|c|}
		\hline
		\textbf{}&\multicolumn{10}{c|}{\textbf{HITS@3}}
		\\\hline
		\textbf{Model Variants}&\textbf{1t}&\textbf{2t}&\textbf{3t}&\boldmath$2\cap$&\boldmath$3\cap$&\boldmath$2\cup$&\boldmath$\cap t$&\boldmath$t\cap$&\boldmath$\cup t$&\textbf{Avg}\\\hline
{\model}-1$t$&.649&.141&.128&.410&.466&.477&.095&.257&.102&.303\\
{\model}-translation&.649&.182&.179&.463&.535&.479&.128&.308&.143&.341\\
{\model}-single&.652&\textbf{.225}&.228&.524&.632&.475&.167&.398&.181&.387\\
{\model}-average&.628&.222&.224&.524&.624&.444&.158&.387&.180&.377\\
{\model}-MLP&.642&\textbf{.225}&.228&\textbf{.526}&.631&.462&.166&.400&.183&.385\\\hline
{\model} (final)&\textbf{.654}&\textbf{.225}&\textbf{.232}&.525&\textbf{.635}&\textbf{.481}&\textbf{.170}&\textbf{.408}&\textbf{.184}&\textbf{.390}\\\hline
	\end{tabular}
	\label{tab:ablation}
\end{table*}
From Table \ref{tab:ablation}, we notice that utilizing only 1$t$ queries significantly reduces the performance of our model by $22.3\%$ and even increasing the scope to all translation queries is still lower in performance by $12.5\%$ for this case. However, we notice that training on all single operator queries results in comparable performance to the final {\model} model. But, given the better overall performance, we utilize all the queries in our final model. For union aggregation, we observe that attention has a clear advantage and both vanilla averaging and MLP lead to a lower performance by $3.33\%$ and $1.28\%$, respectively. Thus, we adopt self-attention in our final model.

\subsection{(RQ3) Case Study: Drug Recommendation}
In this experiment, we utilize the expressive power of {\model}'s query representations to recommend therapeutic drugs for COVID-19 from the DRKG dataset. Drugs in the dataset are already approved for other diseases and the aim is to utilize the drug-protein-disease networks and employ them towards treating COVID-19. This can potentially reduce both the drug development time and cost \cite{pushpakom2019drug}. For this experiment, we utilize the treatment relation in DRKG and retrieve drugs $D: D \xrightarrow[]{treats} X$, where $X$ is a set of SARS diseases related to the COVID-19 virus. Given that we only need these limited set of entity types (only SARS diseases and drugs) and relation types (only treatments), we only consider the DRKG subgraph that contains this necessary set of entities and relations for learning the representations. We compare the recommendations of different models against a set of actual candidates currently in trials for COVID-19. %the candidats are also present in the dataset.\nr{how did you get these actual candidates?}}
We use the top-10 recommendations with the evaluation metrics of precision, recall, and F1-score for comparison.

\begin{table}[htbp]
    \centering
    \caption{Performance comparison of various models on the COVID-19 drug recommendation problem using precision (P), recall (R), and F1-score (F1) metrics. The top three drugs recommended by the models are given in the last column. The recommendations given in \textcolor{ao_english}{green} and \textcolor{red}{red} indicate correct and incorrect predictions, respectively. The last two rows provide the average relative improvement of {\model} compared to the state-of-the-art baselines Q2B and CQD.}
    \begin{tabular}{|l|ccc|c|}
    \hline
    \textbf{Model} & \textbf{P@10} & \textbf{R@10} & \textbf{F1} & \textbf{Top Recommended Drugs} \\
    \hline
    GQE&.119&.174&.141&\textcolor{red}{Piclidenoson}, \textcolor{red}{Ibuprofen}, \textcolor{ao_english}{Chloroquine}\\
BQE&.159&.200&.177&\textcolor{ao_english}{Ribavirin},\textcolor{red}{Oseltamivir}, \textcolor{red}{Ruxolitinib}\\
Q2B&.194&.255&.221&\textcolor{ao_english}{Ribavirin}, \textcolor{ao_english}{Dexamethasone}, \textcolor{red}{Deferoxamine}\\
CQD&.209&.260&.232&\textcolor{ao_english}{Ribavirin}, \textcolor{ao_english}{Dexamethasone}, \textcolor{red}{Tofacitinib}\\
{\model}&\textbf{.217}&\textbf{.269}&\textbf{.251}&\textcolor{ao_english}{Ribavirin}, \textcolor{ao_english}{Dexamethasone}, \textcolor{ao_english}{Hydroxychloroquine}\\\hline
\hline
\textbf{{\model} vs Q2B (\%)}&11.9&5.5&13.6&\\
\textbf{{\model} vs CQD (\%)}&3.8&3.5&8.2&\\
    \hline
    \end{tabular}
    \label{tab:drug_recommendation}
\end{table}
We can observe from Table \ref{tab:drug_recommendation} that {\model} is able to provide the best drug recommendations, across all evaluation metrics. Our model is able to outperform the current methods by atleast $3.8\%$, $3.5\%$, and $8.2\%$ in precision, recall, and F1, respectively. Also, the top recommended drugs by our {\model} are more inline with the current drug development candidates, thus, showing the better performance of our model's query representations.

\begin{figure*}[htbp]
    \centering
    \begin{subfigure}[b]{.71\linewidth}
    \centering
    \includegraphics[width=\linewidth]{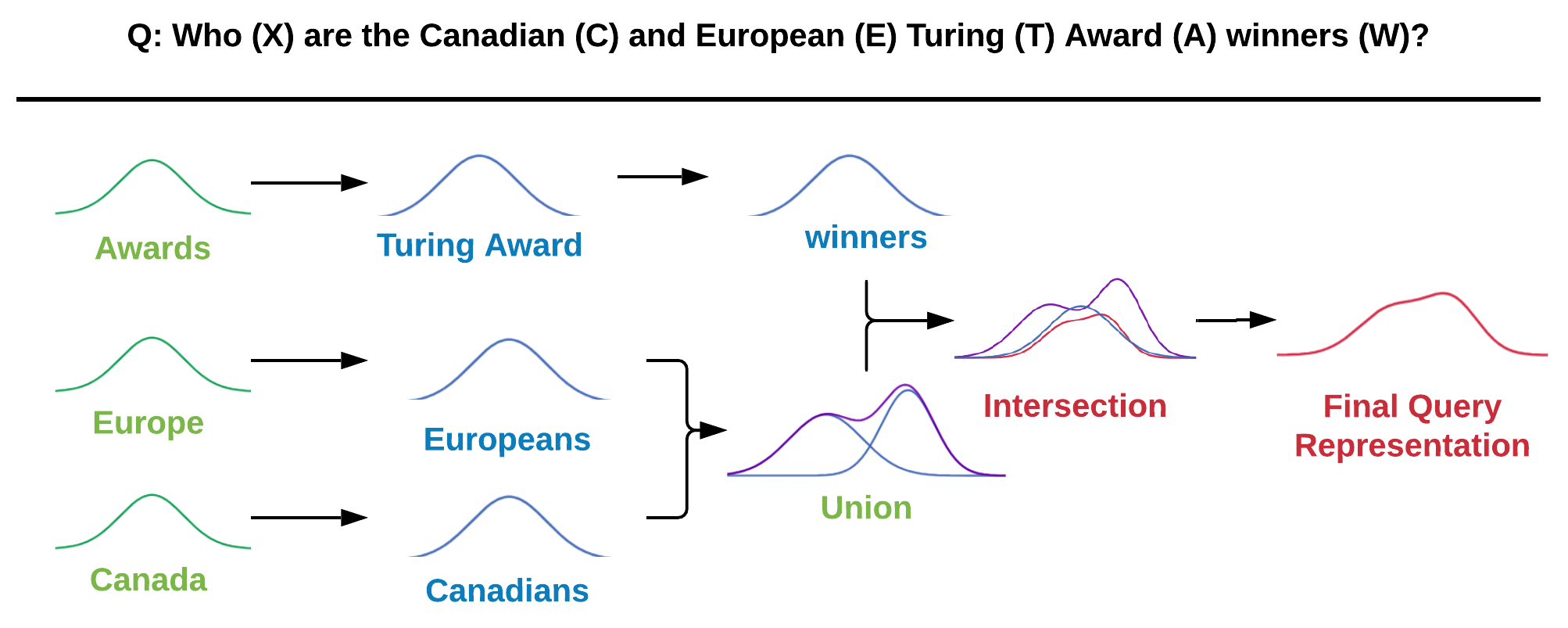}
    \caption{Query processing in {\model}. This figure depicts a univariate version of the entity Gaussian embeddings for better visualization of the process. The same property, however, generalizes over an increased number of dimensions, i.e., multivariate case.}
    \end{subfigure}~~
    \begin{subfigure}[b]{.28\linewidth}
        \centering
        \includegraphics[width=\linewidth]{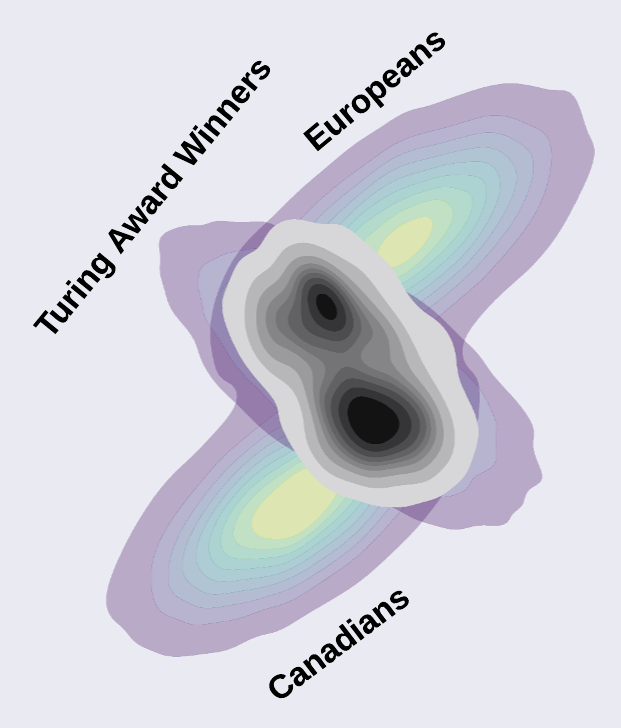}
        \caption{Bivariate version of the final query space, given in grayscale with darker colors representing a higher probability of answers.}
    \end{subfigure}
    \caption{An illustration of the flow for a sample complex query in the representational space. We note that intersection after union is possible in our {\model} model because the operations are closed in {Gaussian distributions} and this is not possible in current methods including BQE, Q2B, and CQD.}
    \label{fig:visualization}
\end{figure*}
\subsection{(RQ4) Visualization of the Gaussian Representations}
To visualize the entity and query in the latent space, we extract representative entity samples from the FB15K-237 dataset and present them in a 2-dimensional space for better comprehension.

Figure \ref{fig:visualization} depicts the different entities and the mechanism through which {\model} narrows down to the particular answer set. Notice that, we are able to perform an intersection after a union operation due to the closed form nature of our operations. This is currently not possible in state-of-the-art baseline methods. Additionally, it should be noted that, unions widen the query space and intersections narrow them down (as expected). Furthermore, the variance parameter acts as a control over the spatial area that an entity should cover and more general entities such as \texttt{Turing Award} and \texttt{Europe} occupy a larger area than their respective sub-categories, namely,  \texttt{winners} and \texttt{Europeans}. %\nr{if time permits, can we add more visualizations in the appendix? for a few other queries}

\section{Conclusion}
\label{sec:conclusion}
In this paper, we present {\fullmodel} ({\model}), a model to learn query representations for chain reasoning over knowledge graphs. We show the {representational power} of \model~by defining {closed form solutions} to FOE queries and their chains. Additionally, we also demonstrate its  superior performance compared to its state-of-the-art counterparts on the problems of reasoning over KGs and drug recommendation for COVID-19 from the DRKG dataset. Furthermore, we exhibit its interpretability by depicting the representational space through a sample query processing pipeline.  

\section{Broader Impact}
\label{sec:broader}
{\model} is the first method that models an individual entity in knowledge graphs using Gaussian density function, making it possible to solve FOE queries using a closed form solution. This enables its application in domains that require chain reasoning. The main idea of the proposed solution can also be extended to any domain that can encode its basic units as Gaussians and extend the units through FOE queries, e.g., in topic modeling, topics can be encoded as Gaussians and documents as union of topics.

However, {\model} depends on the integrity of the  knowledge graph used for training. Any malicious attacks/errors \cite{zugner2018adversarial,dai2018adversarial} that lead to incorrect relations could, further, lead to incorrect results and affect the confidence of our model. Furthermore, due to the connected nature of complex queries, this attack could propagate and affect a larger set of queries. Such incorrect results would be problematic %\nr{fatal is a strong word... maybe "problematic?"} 
in sensitive areas of research such as drug recommendations and cybersecurity and, thus, it is necessary to maintain the integrity of training data before learning representations and querying with {\model}.

% \subsection*{Acknowledgments}
% \textbf{Funding:} This work was supported in part by the US National Science Foundation grant IIS-1838730 and Amazon AWS cloud computing credits for research.

% \textbf{Competing Interests:} All the authors had some form of affiliation with Amazon in the past 36 months. 

\newpage
\bibliographystyle{unsrt}
\bibliography{main}

\newpage
\appendix
The following is the supplementary Appendix for the paper; \textit{Probabilistic Entity Representation Model for Reasoning over Knowledge Graphs}. All the references given in the following sections are made in context of the main paper.

\section{Derivation for Product of Multivariate Gaussians}
\label{app:gaussian_product}
The following sections provide the proof for the product of Gaussians for both the univariate case and multivariate case (used in Eqs.(\ref{eq:q_intersection}) and (\ref{eq:c_intersection})).

\subsection{Univariate Case}
\begin{align*}
    \mathcal{N}(\mu,\sigma) &= \exp\left(\left(\frac{x-\mu}{\sigma}\right)^2\right)\\
    P(\theta) &= P(\theta_1)P(\theta_2) = \exp\left(\left(\frac{x-\mu_1}{\sigma_1}\right)^2\right).\exp\left(\left(\frac{x-\mu_2}{\sigma_2}\right)^2\right)\\
    \log P(\theta) &= \left(\frac{x-\mu_1}{\sigma_1}\right)^2+\left(\frac{x-\mu_2}{\sigma_2}\right)^2\\
    &= \frac{(\sigma_2^2+\sigma_1^2)x^2 - 2(\sigma_1^2\mu_2+\sigma_2^2\mu_1)x + (\mu_1^2\sigma_2^2+\mu_2^2\sigma_1^2)}{\sigma_1^2\sigma_2^2}\\
    % \text{For simplicity, let's say } a=(\sigma_2^2+\sigma_1^2)\text{ and }b=\sigma_1^2\sigma_2^2\\
    % &= \frac{ax^2-2
    &= \frac{x^2 - 2\frac{(\sigma_1^2\mu_2+\sigma_2^2\mu_1)}{\sigma_2^2+\sigma_1^2}x + \frac{\mu_1^2\sigma_2^2+\mu_2^2\sigma_1^2}{{\sigma_2^2+\sigma_1^2}}}{\frac{1}{\sigma_1^2}+\frac{1}{\sigma_2^2}}\\
    &= \left(\frac{x-\frac{(\sigma_1^2\mu_2+\sigma_2^2\mu_1)}{\sigma_2^2+\sigma_1^2}}{(\frac{1}{\sigma_1^2}+\frac{1}{\sigma_2^2})^{-2}}\right)^2+K\text{,where }K = \frac{\mu_1^2\sigma_2^2+\mu_2^2\sigma_1^2}{\sigma_1^2\sigma_2^2}-\left(\frac{\sigma_1^2\mu_2+\sigma_2^2\mu_1}{\sigma_1^2\sigma_2^2}\right)^2\\
    P(\theta) &\propto \exp\left(\left(\frac{x-\frac{(\sigma_1^2\mu_2+\sigma_2^2\mu_1)}{\sigma_2^2+\sigma_1^2}}{(\frac{1}{\sigma_1^2}+\frac{1}{\sigma_2^2})^{-2}}\right)^2\right) \approx \mathcal{N}\left(\frac{(\sigma_1^2\mu_2+\sigma_2^2\mu_1)}{\sigma_2^2+\sigma_1^2},(\frac{1}{\sigma_1^2}+\frac{1}{\sigma_2^2})^{-2}\right)
\end{align*}
\subsection{Multivariate Case}
\begin{align*}
    \mathcal{N}(\mu,\Sigma) &= \exp\left((x-\mu)^T\Sigma^{-1}(x-\mu)\right)\\
    P(\theta) &= P(\theta_1)P(\theta_2) = \exp\left((x-\mu_1)^T\Sigma_1^{-1}(x-\mu_1)\right).\exp\left((x-\mu_2)^T\Sigma_2^{-1}(x-\mu_2)\right)\\
    \log(P(\theta)) &= (x-\mu_1)^T\Sigma_1^{-1}(x-\mu_1) + (x-\mu_2)^T\Sigma_2^{-1}(x-\mu_2)\\
    &= x^T\Sigma_1^{-1}x-\mu_1^T\Sigma_1^{-1}x-x^T\Sigma_1^{-1}\mu_1-\mu_1^T\Sigma_1^{-1}\mu_1+x^T\Sigma_2^{-1}x-\mu_2^T\Sigma_2^{-1}x-x^T\Sigma_2^{-1}\mu_2-\mu_2^T\Sigma_2^{-1}\mu_2\\
    &= x^T(\Sigma_1^{-1}+\Sigma_2^{-1})x -(\mu_1^T\Sigma_1^{-1}+\mu_2^T\Sigma_2^{-1})x -x^T(\Sigma_1^{-1}\mu_1+\Sigma_2^{-1}\mu_2)-(\mu_1^T\Sigma_1^{-1}\mu_1+\mu_2^T\Sigma_2^{-1}\mu_2)\\
    \text{Let's }&\text{assume $P(\theta)\propto \mathcal{N}(\mu_3,\Sigma_3)$, then,}\\
    log(P(\theta)) &= (x-\mu_3)^T\Sigma_3^{-1}(x-\mu_3) + K\\
                   &= x^T\Sigma_3^{-1}x - x^T\Sigma_3^{-1}\mu_3 -\mu_3^T\Sigma_3^{-1}x + \mu_3^T\Sigma_3^{-1}\mu_3+K\\
    \text{Comparing}&\text{ coefficients,}\\
    \Sigma_3^{-1} &= \Sigma_1^{-1}+\Sigma_2^{-1}\\
    \Sigma_3^{-1}\mu_3 &= \Sigma_1^{-1}\mu_1+\Sigma_2^{-1}\mu_2\\
    \implies \mu_3 &= \Sigma_3(\Sigma_1^{-1}\mu_1+\Sigma_2^{-1}\mu_2)\\
    \mu_3&=(\Sigma_1^{-1}+\Sigma_2^{-1})^{-1}(\Sigma_1^{-1}\mu_1+\Sigma_2^{-1}\mu_2)
\end{align*}
Notice that we need $\Sigma_3$ while calculation $\mu_3$. However, to save computational memory, we only store the inverses of covariances, i.e., $\Sigma_1^{-1}, \Sigma_2^{-1} \text{ and } \Sigma_3^{-1}$. So, to solve for $\mu_3$ and avoid the computationally expensive process of matrix inversion, we use the linear solver \texttt{torch.solve} on the equation $\Sigma_3^{-1}\mu_3 = \Sigma_1^{-1}\mu_1+\Sigma_2^{-1}\mu_2$.

\section{Algorithm for KG Reasoning with {\model}}
\label{app:algorithm}
Algorithm \ref{alg:PERM} provides an outline of {\model}'s overall framework to learn representations of entities $e \in E$ and relations $r \in R$. The algorithm describes the training from FOE operations of translation (lines \ref{alg:start_translation}-\ref{alg:end_translation}), intersection (lines \ref{alg:start_intersection}-\ref{alg:end_intersection}), and union (lines \ref{alg:start_union}-\ref{alg:end_union}).

\begin{algorithm}[htbp]
	\SetAlgoLined
	\KwIn{Training data $D_t, D_\cap, D_\cup$, which are set of all (query ($Q$), result ($V$)) for translation, intersection, and union, respectively;}
	\KwOut{ Entity $E$ and Relation $R$ gaussian density functions; }
	Randomly initialize $e=\mathcal{N}(\mu_e,\Sigma_e) \in E$ and $r=\mathcal{N}(\mu_r,\Sigma_r) \in R$);\\
	\For{number of epochs; until convergence}
	{
	 $l = 0$;{\color{blue}{ \# Initialize loss}\\}
	\For{$\{(e,r, V_t) \in D_t\}$} 
	{
		\label{alg:start_translation}
		$q_t = \mathcal{N}(\mu_e+\mu_r,(\Sigma_e^{-1}+\Sigma_r^{-1})^{-1})$ from Eq. \eqref{eq:translation}\\
		\nonl {\color{blue}{ \# Update loss for translation queries}\\}
		\oldnl$l = l + \sum_{v_t \in V_t}d_{\mathcal{N}}(v_t,q_t)$
	}\label{alg:end_translation}
	\For{$\{(Q_\cap, V_\cap) \in D_\cap\}$} 
	{	
		\label{alg:start_intersection}
		$q_\cap= \mathcal{N}(\mu_3,\Sigma_3)$, from Eq. \eqref{eq:q_intersection}\\
		\nonl {\color{blue}{ \# Update loss for intersection queries}\\}
		\oldnl$l = l + \sum_{v_\cap \in V_\cap}d_{\mathcal{N}}(v_\cap,q_\cap)$
	} \label{alg:end_intersection}
	\For{$\{(Q_\cup, V_\cup) \in D_\cup\}$}
	{	
		 \label{alg:start_union}
		$q_\cup= \sum_{i=1}^n\phi_i\mathcal{N}(\mu_{e_i},\Sigma_{e_i})$ from Eq. \eqref{eq:q_union}\\
		 \nonl{\color{blue}{ \# Update loss for union queries}\\}
		\oldnl$l = l + \sum_{v_\cup \in V_\cup}\sum_{i=1}^n\phi_id_\mathcal{N}(v_\cup,\mathcal{N}(\mu_{e_i},\Sigma_{e_i}))$
	} \label{alg:end_union}
	%\If{$!(l \approx 0)$}
	{
		\nonl{\color{blue}{\# Update E and R with backpropagation}\\}
		\oldnl$E \leftarrow E - \Delta_E l$\\
		\oldnl$R \leftarrow R - \Delta_R l$
	}
	}
	\Return{E,R}
	\caption{{\model} training algorithm}
	\label{alg:PERM}
\end{algorithm}

\section{MRR metrics for Reasoning over KGs}
\label{app:mrr}
Table \ref{tab:query_mrr} provides the Mean Reciprocal Rank (MRR) results for the reasoning over KGs experiment, given in section \ref{sec:exp}.
\begin{table*}[htbp]
	\centering
	\caption{Performance comparison of {\model} against the baselines to study the efficacy of the query representations. The columns present the different query structures and the overall average performance. The last two rows presents the Average Relative Improvement (\%) of {\model} compared to Q2B and CQD over all datasets across query types. Best results for each dataset are shown in bold.}
	\begin{tabular}{|ll|ccc|ccc|ccc|c|}
		\hline
		\textbf{Metrics}&&\multicolumn{10}{c|}{\textbf{Mean Reciprocal Rank}}
		\\\hline
		\textbf{Dataset}&\textbf{Model}&\textbf{1t}&\textbf{2t}&\textbf{3t}&\boldmath$2\cap$&\boldmath$3\cap$&\boldmath$2\cup$&\boldmath$\cap t$&\boldmath$t\cap$&\boldmath$\cup t$&\textbf{Avg}\\\hline
FB15k-237&GQE&.346&.191&.144&.258&.361&.144&.087&.164&.149&.205\\
&BQE&.390&.109&.100&.228&\textbf{.425}&.124&.224&.126&.097&.203\\
&Q2B&.400&.225&.173&.275&.378&.198&.105&.180&.178&.235\\
&CQD&.439&\textbf{.270}&\textbf{.206}&.299&.381&.235&\textbf{.271}&\textbf{.415}&.112&.292\\
&{\model}&\textbf{.445}&.268&.201&\textbf{.306}&.409&\textbf{.253}&.269&.353&\textbf{.220}&\textbf{.303}\\\hline
NELL995&GQE&.311&.193&.175&.273&.399&.159&.078&.168&.130&.210\\
&BQE&\textbf{.530}&.130&.114&\textbf{.376}&\textbf{.475}&.122&\textbf{.241}&.143&.085&.246\\
&Q2B&.413&.227&.208&.288&.414&.266&.125&.193&.155&.254\\
&CQD&.442&\textbf{.251}&\textbf{.226}&.304&.441&\textbf{.348}&.124&\textbf{.212}&.104&\textbf{.273}\\
&{\model}&.432&.244&.217&.296&.438&.332&.122&.178&\textbf{.190}&.272\\\hline
DBPedia&GQE&.502&.005&N.A.&.749&.773&.320&.154&.597&0.00&.388\\
&BQE&.657&\textbf{.006}&N.A.&\textbf{.964}&\textbf{.966}&.306&\textbf{.419}&.527&0.00&.481\\
&Q2B&.619&\textbf{.006}&N.A.&.840&.863&.468&.212&.779&0.00&.473\\
&CQD&.648&\textbf{.006}&N.A.&.840&.863&.485&.206&.716&0.00&.471\\
&{\model}&\textbf{.706}&\textbf{.006}&N.A.&.841&.862&\textbf{.564}&.219&\textbf{.869}&0.00&\textbf{.508}\\\hline
DRKG&GQE&.313&.182&.132&.232&.360&.144&.097&.166&.163&.199\\
&BQE&.413&.118&.106&.298&.451&.147&\textbf{.270}&.154&.116&.230\\
&Q2B&.371&.225&.178&.283&.422&.205&.064&.122&.223&.233\\
&CQD&.413&\textbf{.277}&\textbf{.213}&.310&.427&.246&.174&\textbf{.282}&.143&.276\\
&{\model}&\textbf{.420}&.276&.211&\textbf{.325}&\textbf{.465}&\textbf{.271}&.179&.249&\textbf{.282}&\textbf{.298}\\\hline
\multicolumn{2}{|c|}{\textbf{{\model} vs Q2B (\%)}}&11.1&16.3&12.5&4.90&4.70&24.9&55.9&29.4&24.5&20.5\\
\multicolumn{2}{|c|}{\textbf{{\model} vs CQD (\%)}}&4.20&-1.2&-2.5&1.40&4.40&10.6&1.80&1.50&92.8&12.6\\\hline
\end{tabular}
	\label{tab:query_mrr}
\end{table*}
\begin{table*}[htbp]
\footnotesize
	\centering
	\caption{Performance comparison of (final) {\model} model against its variants to study the contributions of its components. The columns present the query structures and the overall average performance.}
	\begin{tabular}{|l|l|ccc|ccc|ccc|c|}
		\hline
		\textbf{Metrics}&&\multicolumn{10}{c|}{\textbf{HITS@3}}
		\\\hline
		\textbf{Dataset}&\textbf{Variants}&\textbf{1t}&\textbf{2t}&\textbf{3t}&\boldmath$2\cap$&\boldmath$3\cap$&\boldmath$2\cup$&\boldmath$\cap t$&\boldmath$t\cap$&\boldmath$\cup t$&\textbf{Avg}\\\hline
FB15k-237&1t&.516&.179&.119&.282&.360&.302&.071&.134&.133&.233\\
&translations&.516&.231&.167&.318&.413&.304&.096&.160&.185&.266\\
&single&.511&.282&.212&.359&.486&.296&.126&.207&.235&.302\\
&average&.499&.282&.209&.360&.482&.282&.119&.201&.234&.296\\
&MLP&.510&.285&.212&.363&.488&.293&.125&.208&.238&.302\\
&(final)&.520&.286&.216&.361&.490&.305&.128&.212&.239&.306\\\hline
NELL995&1t&.576&.179&.134&.275&.373&.456&.072&.123&.111&.255\\
&translations&.576&.231&.188&.310&.428&.458&.097&.147&.155&.288\\
&single&.571&.282&.239&.350&.504&.446&.127&.190&.197&.323\\
&average&.558&.282&.235&.351&.500&.425&.120&.185&.196&.317\\
&MLP&.570&.285&.239&.354&.506&.442&.126&.191&.199&.324\\
&(final)&.581&.286&.243&.352&.508&.460&.129&.195&.200&.328\\\hline
DBPedia&1t&.942&.004&N.A.&.781&.734&.775&.129&.600&0.00&.496\\
&translations&.942&.006&N.A.&.881&.843&.779&.174&.718&0.00&.543\\
&single&.934&.007&N.A.&1.00&1.00&.758&.228&.928&0.00&.607\\
&average&.912&.007&N.A.&.997&.984&.723&.216&.903&0.00&.593\\
&MLP&.932&.007&N.A.&.996&.992&.751&.227&.932&0.00&.605\\
&(final)&.950&.007&N.A.&1.00&1.00&.782&.232&.952&0.00&.615\\\hline
DRKG&1t&.560&.202&.130&.302&.396&.373&.106&.172&.165&.267\\
&translations&.560&.260&.183&.341&.455&.374&.143&.206&.230&.306\\
&single&.555&.317&.232&.385&.536&.365&.187&.266&.293&.348\\
&average&.543&.317&.228&.386&.531&.347&.177&.259&.291&.342\\
&MLP&.554&.321&.232&.389&.538&.361&.186&.267&.296&.349\\
&(final)&.565&.322&.236&.387&.540&.376&.190&.273&.297&.354\\\hline
\end{tabular}
\begin{tabular}{|l|l|ccc|ccc|ccc|c|}
		\hline
		\textbf{Metrics}&&\multicolumn{10}{c|}{\textbf{Mean Reciprocal Rank}}
		\\\hline
		\textbf{Dataset}&\textbf{Variants}&\textbf{1t}&\textbf{2t}&\textbf{3t}&\boldmath$2\cap$&\boldmath$3\cap$&\boldmath$2\cup$&\boldmath$\cap t$&\boldmath$t\cap$&\boldmath$\cup t$&\textbf{Avg}\\\hline
FB15k-237&{\model}-1t&.410&.180&.122&.217&.274&.209&.085&.127&.145&.197\\
&translations&.410&.232&.171&.245&.314&.210&.115&.152&.202&.228\\
&single&.406&.283&.217&.277&.370&.204&.151&.197&.257&.262\\
&average&.396&.283&.214&.278&.367&.194&.143&.191&.256&.258\\
&MLP&.405&.286&.217&.280&.372&.202&.150&.198&.260&.263\\
&(final)&.445&.268&.201&.306&.409&.253&.269&.353&.220&.303\\\hline
NELL995&1t&.432&.191&.160&.234&.275&.332&.094&.162&.125&.223\\
&translations&.428&.197&.168&.261&.369&.331&.092&.134&.147&.236\\
&single&.425&.241&.213&.294&.435&.322&.120&.173&.187&.268\\
&average&.415&.241&.210&.295&.431&.307&.113&.169&.186&.263\\
&MLP&.424&.243&.213&.298&.436&.319&.119&.174&.189&.268\\
&(final)&.432&.244&.217&.296&.438&.332&.122&.178&.190&.272\\\hline
DBPedia&1t&.706&.005&N.A.&.665&.541&.564&.169&.791&0.00&.430\\
&translations&.700&.005&N.A.&.741&.727&.562&.164&.655&0.00&.444\\
&single&.694&.006&N.A.&.841&.862&.547&.215&.847&0.00&.502\\
&average&.678&.006&N.A.&.838&.848&.521&.204&.824&0.00&.490\\
&MLP&.693&.006&N.A.&.838&.855&.542&.214&.851&0.00&.500\\
&(final)&.706&.006&N.A.&.841&.862&.564&.219&.869&0.00&.452\\\hline
DRKG&1t&.416&.173&.116&.254&.341&.269&.100&.157&.157&.220\\
&translations&.416&.223&.164&.286&.392&.270&.135&.188&.218&.255\\
&single&.413&.272&.207&.323&.462&.263&.176&.243&.278&.293\\
&average&.404&.272&.204&.324&.457&.250&.167&.236&.276&.288\\
&MLP&.412&.275&.207&.327&.463&.260&.175&.244&.281&.294\\
&(final)&.420&.276&.211&.325&.465&.271&.179&.249&.282&.298\\\hline
\end{tabular}
	\label{tab:finer_ablation}
\end{table*}
\section{Finer Evaluation of Ablation Study}
\label{app:ablation}
Table \ref{tab:finer_ablation} provides finer results of our ablation study.

\end{document}